\documentclass[11pt,a4paper]{article}
\usepackage[hyperref]{acl2021}
\usepackage{times}
\pdfoutput=1
\usepackage{latexsym}
\usepackage{amssymb}
\usepackage{amsmath} 
\usepackage{booktabs}
\usepackage{enumerate}
\usepackage{graphicx}
\usepackage{subfigure}
\usepackage{xspace}
\usepackage{float}
\usepackage{bbm}
\usepackage{bm}
\usepackage{multirow}
\usepackage{booktabs}
\usepackage{color}
\usepackage{framed}
\usepackage{stfloats}
\usepackage{iitem}
\usepackage[T1]{fontenc}
\definecolor{shadecolor}{RGB}{180,180,180}
\usepackage{url}
\newcommand{\paratitle}[1]{\vspace{1.5ex}\noindent\textbf{#1}}
\newcommand{\ie}{\emph{i.e.,}\xspace}

\newcommand{\eg}{\emph{e.g.,}\xspace}

\newcommand{\ignore}[1]{}

\usepackage{microtype}
\usepackage{algorithm}

\usepackage{algpseudocode}
\usepackage{amsthm}
\usepackage{amsmath}
\usepackage{tikz}

\usepackage{amsmath}

\newcommand\Vector{\bm}
\newcommand\Matrix{\mathbf}            
\newcommand\Tensor{\mathcal}

\aclfinalcopy 
\def\aclpaperid{***} 

\date{}

\begin{document}

\title{Enabling Lightweight Fine-tuning for Pre-trained Language Model Compression based on Matrix Product Operators}


\author{
	Peiyu Liu$^{1,4}$\thanks{$\ $ Authors contributed equally.}\ \ ,
	Ze-Feng Gao$^{2,1*}$\ ,
	Wayne Xin Zhao$^{1,4,5}$\thanks{$\ $ Corresponding author.}\ \ ,
	\\ \textbf{Z.Y. Xie}$^{2}$\textbf{,}
	\textbf{Zhong-Yi Lu}$^{2\dagger}$\and
	\textbf{Ji-Rong Wen}$^{1,3,4}$
	\\
	$^1$Gaoling School of Artificial Intelligence, Renmin University of China\\
	$^2$Department of Physics, Renmin University of China\\
	$^3$ School of Information, Renmin University of China\\
	$^4$Beijing Key Laboratory of Big Data Management and Analysis Methods\\
	$^5$Beijing Academy of Artificial Intelligence, Beijing, 100084, China\\
	{\{liupeiyustu,zfgao,qingtaoxie,zlu,jrwen\}@ruc.edu.cn, batmanfly@gmail.com}\\ 
}

\maketitle

\begin{abstract}
This paper presents a novel pre-trained language models (PLM) compression approach based on the matrix product operator (short as MPO) from quantum many-body physics.
It can decompose an original matrix into central tensors (containing the core information) and auxiliary tensors (with only a small proportion of parameters). 
With the decomposed MPO structure, we propose a novel fine-tuning strategy by only updating the parameters from the auxiliary tensors, and design an optimization algorithm for MPO-based approximation over stacked network architectures. %
Our approach can be applied to the original or the compressed PLMs in a general way, which derives a lighter network and significantly reduces the parameters to be fine-tuned. 
Extensive experiments have demonstrated the effectiveness of the proposed approach in model compression, especially the reduction in fine-tuning parameters (91$\%$ reduction on average). 
The code to reproduce the results of this paper can be found at \url{https://github.com/RUCAIBox/MPOP}.


\end{abstract}

\section{Introduction}
Recently, pre-trained language models~(PLMs)~\cite{devlin2018bert,peters2018deep,radford2018improving} have made significant progress in various natural language processing tasks. Instead of training a model from scratch, one can fine-tune a PLM to solve some specific task through the paradigm of ``\emph{pre-training} and \emph{fine-tuning}''.

Typically, PLMs are constructed with stacked Transformer layers~\cite{vaswani2017attention}, involving a huge number of parameters to be learned. Though effective, the large model size makes it impractical for resource-limited devices. Therefore, there is an increasing number of studies focused on the parameter reduction or memory reduction of PLMs~\cite{noach2020compressing}, including parameter sharing~\cite{lan2019albert}, knowledge distillation~\cite{sanh2019distilbert}, low-rank approximation~\cite{ma2019tensorized} and data quantization~\cite{hubara2017quantized}.
However, these studies mainly apply parameter reduction techniques to PLM compression, which may not be intrinsically appropriate for the learning paradigm and architecture of PLMs.
The compressed parameters are highly coupled so that it is difficult to directly manipulate different parts with specific strategies.
For example, most PLM compression methods need to fine-tune the whole network architecture, although only a small proportion of parameters will significantly change during fine-tuning~\cite{liu2020fastbert}. 

In this paper, we introduce a novel matrix product operator~(MPO) technique from quantum many-body physics for compressing PLMs~\cite{gao2020compressing}. The MPO is an algorithm that factorizes a matrix into a sequential product of local tensors (\ie a multi way array). Here, we call the tensor right in the middle as \emph{central tensor} and the rest as \emph{auxiliary tensors}. An important merit of the MPO decomposition is structural in terms of information distribution: the central tensor with most of the parameters encode the core information of the original matrix, while the auxiliary tensors with only a small proportion of parameters play the role of complementing the central tensor. Such a property motivates us to investigate whether such an MPO can be applied to derive a better PLM compression approach: can we compress the central tensor for parameter reduction and update the auxiliary tensors for lightweight fine-tuning? If this could be achieved, we can derive a lighter network meanwhile reduce the parameters to be fine-tuned.

To this end, we propose an \underline{MPO}-based compression approach for \underline{P}LMs, called \emph{MPOP}. It is developed based on the MPO decomposition technique~\cite{gao2020compressing,pirvu2010matrix}. 
We have made two critical technical contributions for compressing PLMs with MPO. First, we introduce a new fine-tuning strategy that only focuses on the parameters of auxiliary tensors, so the number of fine-tuning parameters can be largely reduced. We present both theoretical analysis and experimental verification for the effectiveness of the proposed fine-tuning strategy.
Second, we propose a new optimization algorithm, called \emph{dimension squeezing}, tailored for stacked neural layers.
Since mainstream PLMs usually consist of multiple Transformer layers, this will produce accumulated reconstruction error by directly applying low-rank approximation with MPO at each layer. 
The dimension squeezing algorithm is able to gradually perform the dimension truncation in a more stable way so that it can dramatically alleviate the accumulation error in the stacked architecture.

To our knowledge, it is the first time that MPO is applied to the PLM compression, which is well suited for both the learning paradigm and the architecture  of PLMs.  We construct  experiments to evaluate the effectiveness of the proposed compression approach for ALBERT, BERT, DistillBERT and MobileBERT, respectively, on GLUE benchmark. Extensive experiments have demonstrated the effectiveness of the proposed approach in model compression, especially dramatically reducing the fine-tuning parameters (91$\%$ reduction on average).
\section{Related Work}
We review the related works in three aspects.

\paratitle{Pre-trained Language Model Compression}.
Since the advent of large-scale PLMs, several variants were proposed to alleviate its memory consumption. For example, DistilBERT~\cite{sanh2019distilbert} and MobileBERT~\cite{sun2020mobilebert} leveraged knowledge distillation to
reduce the BERT network size. 
SqueezeBERT~\cite{iandola2020squeezebert} and Q8BERT~\cite{zafrir2019q8bert} adopted special techniques to substitute the operations or quantize both weights and activations. 
ALBERT~\cite{lan2019albert} introduced cross-layer parameter sharing and low-rank approximation to reduce the number of parameters. More studies~\cite{jiao2019tinybert,hou2020dynabert,liu2020fastbert,wang2020linformer,khetan-karnin-2020-schubert,xin2020deebert,pappas2020grounded,sun2020contrastive} can be found in the comprehensive survey~\cite{ganesh2020compressing}.

\paratitle{Tensor-based Network Compression}. Tensor-based methods have been successfully applied to neural network compression.
For example, MPO has been utilized to compress linear layers of deep neural network~\cite{gao2020compressing}.  \citet{sun2020model} used MPO to compress the LSTM model on acoustic data.  \citet{novikov2015tensorizing} coined the idea of reshaping weights of fully-connected layers into high-dimensional tensors and representing them in Tensor Train~(TT)~\cite{oseledets2011tensor} format, which was extended to other network architectures~\cite{garipov2016ultimate,yu2017long,tjandra2017compressing,khrulkov2019tensorized}.  \citet{ma2019tensorized} adopted block-term tensor decomposition to compress Transformer layers in PLMs.

\paratitle{Lightweight Fine-tuning.}  In the past, lightweight fine-tuning was performed without considering parameter compression.
As a typical approach, trainable modules are inserted into PLMs. For example,  a ``side'' network is fused with PLM via summation in \citep{zhang2019bertscore}, and adapter-tuning inserts task-specific layers (adapters) between each layer of PLMs~\cite{houlsby2019parameter,lin2020exploring,rebuffi2017learning}.
On the contrary, several studies consider removing parameters from PLMs. For example, several model weights are ablated away by training a binary parameter mask~\cite{zhao2020masking,radiya2020fine}.

Our work is highly built on these studies, while we have a new perspective by designing the PLM compression algorithm, which enables lightweight fine-tuning. It is the first time that MPO is applied to PLM compression, and we make two major technical contributions for achieving lightweight fine-tuning and stable optimization.

\section{Preliminary}\label{sec-preliminary}

In this paper, scalars are denoted by lowercase letters  (\eg $a$), vectors are denoted by boldface lowercase letters  (\eg $\Vector{v}$), matrices are denoted by boldface capital letters (\eg $\Matrix{M}$), and high-order (order three or higher) tensors are denoted by boldface Euler script letters (\eg $\Tensor{T}$). An $n$-order tensor $\Tensor{T}_{i_1,i_2,...i_n}$ can be considered as a multidimensional array with $n$ indices $\{ i_1,i_2,...,i_n \}$.

\begin{figure*}
    \centering
    \includegraphics[width=0.9\textwidth]{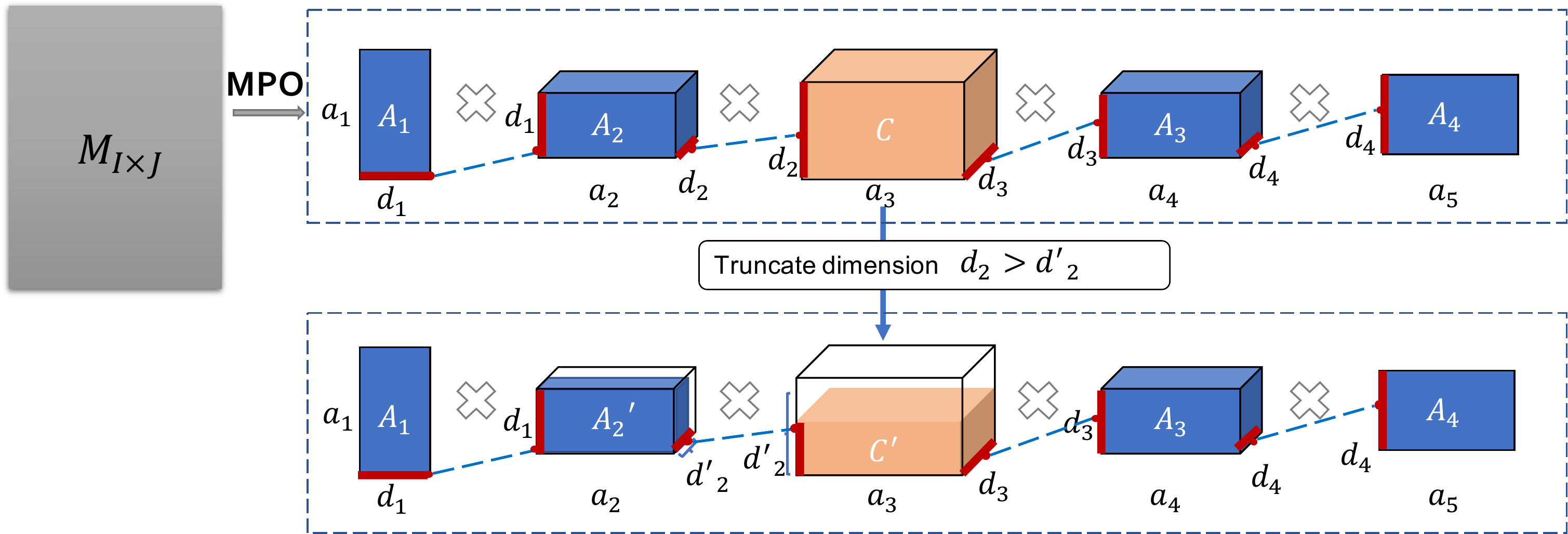}
    \caption{MPO decomposition for matrix $\Matrix{M}_{I\times J}$ with five local tensors, where $\prod_{k=1}^{n}i_k=I, \prod_{k=1}^{n}j_k=J$, and $a_k = i_k \times j_k$~($n=5$ here).
    Auxiliary tensors~($\{\Tensor{A}_{i}\}_{i=1}^{4}$) and central tensor~($\Tensor{C}$) are marked in blue and orange, respectively. Dash line linking adjacent tensors denotes virtual bonds.
    }
    \label{fig:MPO-decomposition}
\end{figure*}

\paratitle{Matrix Product Operator}. Originating from quantum many-body physics, 
matrix product operator~(MPO) is a standard algorithm to factorize a matrix into a sequential product of multiple local tensors~\cite{gao2020compressing, pirvu2010matrix}. 
Formally, given a matrix $\Matrix{M}\in \mathbb{R}^{I\times J}$, 
its MPO decomposition into a product of $n$ local tensors can be represented as:
\begin{equation}
    \textsc{MPO}~(\Matrix{M})=\prod_{k=1}^{n} \Tensor{T}_{(k)}[d_{k-1},i_k,j_k,d_k],
    \label{eq:mpo}
\end{equation}
where the $\Tensor{T}_{(k)}[d_{k-1},i_k,j_k,d_k]$ is a 4-order tensor with size $d_{k-1}\times i_k \times j_k \times d_k$ in which $\prod_{k=1}^{n}i_k=I, \prod_{k=1}^{n}j_k=J$ and $d_0=d_n=1$.
We use the concept of \emph{bond} to connect two adjacent tensors~\cite{pirvu2010matrix}. The bond dimension $d_k$ is defined by:
\begin{equation}
    d_k = \min\bigg(\prod_{m=1}^k i_m\times j_m, \prod_{m=k+1}^n i_m\times j_m\bigg).
    \label{eq:d-k}
\end{equation}
From Eq.~\eqref{eq:d-k}, we can see that $d_k$ is going to be large in the middle and small on both sides. We present a detailed algorithm for MPO decomposition in Algorithm~\ref{alg:mpo-decomposition}.
In this case, we refer to the tensor right in the middle as \emph{central tensor}, and the rest as \emph{auxiliary tensor}. Figure~\ref{fig:MPO-decomposition} presents the illustration of MPO decomposition, and we use $n=5$ in this paper.

\floatname{algorithm}{Algorithm}
\begin{algorithm}
\small
    \caption{MPO decomposition for a matrix.}   
    \begin{algorithmic}[1] 
        \Require matrix $\Matrix{M}$, the number of local tensors $n$
        \Ensure: MPO tensor list $ \{\Tensor{T}_{(k)}\}_{k=1}^{n}$
        \For{$k=1 \to n-1$ }
            \State $\Matrix{M}[I,J]\longrightarrow \Matrix{M}[d_{k-1}\times i_k\times j_k,-1]$
            \State $\Matrix{U}\lambda \Matrix{V}^\top=\rm{SVD}~(\Matrix{M})$
            \State $\Matrix{U}[d_{k-1}\times i_k\times j_k, d_k]\longrightarrow \Tensor{U}[d_{k-1},i_k,j_k,d_k]$
            \State $\Tensor{T}^{(k)}:= \Tensor{U} $
            \State $\Matrix{M}:=\lambda \Matrix{V}^{\top}$
        \EndFor
        \State $\Tensor{T}^{(n)}:=\Matrix{M}$
        \State $\rm{Normalization}$
        \State \Return{$ \{\Tensor{T}_{(k)}\}_{k=1}^{n}$}
    \end{algorithmic}
\label{alg:mpo-decomposition}
\end{algorithm}

\paratitle{MPO-based Low-Rank Approximation}.
With the standard MPO decomposition in Eq.~\eqref{eq:mpo}, we can exactly reconstruct the original matrix $\mathbf{M}$ through the product of the derived local tensors.
Following~\cite{gao2020compressing}, we can truncate the $k$-th bond dimension $d_k$  (see Eq.~\eqref{eq:mpo}) of local tensors to $d'_k$  for low-rank approximation:  $d_k > d'_k$. We can set different values for $\{d_k\}_{k=1}^n$ to control the expressive capacity of  MPO-based reconstruction.  The truncation error induced by the $k$-th bond dimension $d_k$ is denoted by $\epsilon_k$ (called \emph{local truncation error}) which can be efficiently computed as:
\begin{equation}
    \epsilon_k = \sum_{i=d_k-d'_k}^{d_k}\lambda_i, \\
    \label{eq:epsilion-k}
\end{equation}
where $\{\lambda_i\}_{i=1}^{d_k}$ are the singular values of $\Matrix{M}[i_1j_1...i_k j_k, i_{k+1}j_{k+1}...i_n j_n]$.

Then the total truncation error satisfies:
\begin{equation}
    \left \| \Matrix{M}-\textsc{MPO}(\Matrix{M}) \right \|_F \leq \sqrt{\sum_{k=1}^{n-1} \epsilon_k^2}.
    \label{eq:mpo-truncate-error}
\end{equation}
The proof can be found in the supplementary materials~\footnote{https://github.com/RUCAIBox/MPOP}.
Eq.~\eqref{eq:mpo} indicates that the reconstruction error is bounded by the sum of the squared local truncation errors, which is easy to estimate in practice.

Suppose that we have truncated the dimensions of local tensors from $\{d_{k}\}_{k=1}^{n}$ to $\{d'_{k}\}_{k=1}^{n}$, the compression ratio introduced by quantum many-body physics~\cite{gao2020compressing} can be computed as follows:
\begin{equation}
    \rho = \frac{\sum_{k=1}^{n} d'_{k-1}i_k j_k  d'_k}{\prod_{k=1}^{n} i_k j_k }.
    \label{eq:mpo-compression-rate}
\end{equation}
The smaller the compression ratio is, the fewer parameters are kept in the MPO representation. On the contrary, the larger the compression ratio $\rho$ is, and the more parameters there are, and the smaller the reconstruction error is. 
When $\rho>1$, it indicates the decomposed tensors have more parameters than the original matrix.

\section{Approach}

So far, most of pre-trained language models~(PLM) are developed based on stacked Transformer layers~\cite{vaswani2017attention}.
Based on such an architecture, it has become a paradigm to first pre-train PLMs and then fine-tunes them on task-specific data. 
The involved parameters of PLMs can be generally represented in the matrix format. Hence, it would be natural to apply MPO-based approximation for compressing the parameter matrices in PLMs by truncating tensor dimensions.

In particular, we propose two major improvements for MPO-based PLM compression, which can largely reduce the fine-tuning parameters and effectively improve the optimization of stacked architecture, respectively.  

\begin{table}[]
\small
\centering
\begin{tabular}{llll}
\toprule[1pt]
Layers           & (0,1e-4] & (1e-4,1e-3] &(1e-3,$\infty$) \\ \midrule[0.7pt]
Word embedding   & 0.66     &  0.26    &0.09\\
Feed-forward     & 0.09     &  0.64    &0.27\\
Self-attention   & 0.09     &  0.64    &0.27\\
\bottomrule[1pt]
\end{tabular}
\caption{Distribution of parameter variations for BERT when fine-tuned  on SST-2 task.}
\label{tab-absolute_err}
\end{table}

\subsection{Lightweight Fine-tuning with Auxiliary Tensors}
\label{sec-auxiliary}
Due to the high coupling of parameters, previous PLM compression methods usually need to fine-tune all the parameters. 
As a comparison, the MPO approach decomposes a matrix into a list of local tensors, which makes it potentially possible to consider fine-tuning different parts with specific strategies. 
Next, we study how to perform lightweight fine-tuning based on MPO properties. 

\paratitle{Parameter Variation from Pre-Training}. 
To apply our solution to lightweight fine-tuning, we first conduct an empirical experiment to check the variation degree of the parameters \emph{before} and \emph{after} fine-tuning. Here, we adopt the standard pre-trained BERT~\cite{devlin2018bert} and then fine-tune it on the SST-2 task~\cite{socher2013recursive}. We first compute the absolute difference of the variation for each parameter value and then compute the ratio of parameters with different variation levels. The statistical results are reported in Table~\ref{tab-absolute_err}. As we can see, most of parameters vary little, especially for the word embedding layer. This finding has also been reported in a previous studies~\cite{khetan-karnin-2020-schubert}. 
As discussed in Section~\ref{sec-preliminary}, after MPO decomposition, the central tensor contains the majority of the parameters, while the auxiliary tensors only contain a small proportion of the parameters. Such merit inspires us to consider only fine-tuning the parameters in the auxiliary tensors while keeping the central tensor fixed during fine-tuning.  
If this approach was feasible, this will largely reduce the parameters to be fine-tuned. 

\paratitle{Theoretical Analysis}. Here we introduce entanglement entropy from  quantum mechanics~\cite{calabrese2004entanglement} as the metric to measure the information contained in MPO bonds, which is similar to the entropy in information theory but replaces probabilities by normalized singular values produced by SVD. This will be more suitable for measuring the information of a matrix as singular values often correspond to the important information implicitly encoded in the matrix, and the importance is positively correlated with the magnitude of the singular values. Following~\cite{calabrese2004entanglement}, the entanglement entropy $S_k$ corresponding to the $k$-th bond can be calculated by:
\begin{equation}
    S_{k} = -\sum_{j=1}^{d_k}v_{j}\ln v_{j}, \quad k = 1,2,...,n-1,
    \label{eq:entropy}
\end{equation}
where $\{v_{j}\}_{j=1}^{d_k}$ denote the normalized SVD eigenvalues of $\Matrix{M}[i_1j_1...i_k j_k,i_{k+1}j_{k+1}...i_n j_n]$.
The entanglement entropy $S_{k}$ is an increasing function of dimension $d_k$ as described in~\cite{gao2020compressing}. Based on Eq.~\eqref{eq:d-k},  the central tensor has the largest bond dimension, corresponding to the largest entanglement entropy. This indicates that most of the information in an original matrix will be concentrated in the central tensor. Furthermore, the larger a dimension is, the larger the updating effect will be.  According to \cite{pirvu2010matrix}, it is also guaranteed in principle that any change on some tensor will be transmitted to the whole local tensor set. Thus, it would have almost the same effect after convergence by optimizing the central tensor or the auxiliary tensors for PLMs. 

Based on the above analysis, we speculate that the affected information during fine-tuning is mainly encoded on the auxiliary tensors so that the overall variations are small. Therefore, for lightweight fine-tuning, we first perform the MPO decomposition for a parameter matrix, and then only update its auxiliary tensors according to the downstream task with the central tensor fixed.
Experimental results in Section~\ref{sec:experimental-results} will demonstrate that such an approach is indeed effective.

\subsection{Dimension Squeezing for Stacked Architecture  Optimization}
\label{sec-squeeze}
Most of PLMs are stacked with multiple Transformer layers. 
Hence, a major problem with directly applying MPO to compressing PLMs is that the reconstruction error tends to be accumulated and amplified exponentially by the number of layers. It is thus urgent to develop a more stable optimization algorithm tailored to the stacked architecture. 

\paratitle{Fast Reconstruction Error Estimation}. Without loss of generality, we can consider a simple case in which each layer contains exactly one parameter matrix to be compressed. Assume that there are $L$ layers, so we have $L$ parameter matrices in total, denoted by $\{ \Matrix{M}^{(l)} \}_{l=1}^L$. Let $\Tensor{C}^{(l)}$ denote the corresponding central tensor with a specific dimension $d^{(l)}$ after decomposing $\Matrix{M}^{(l)}$ with MPO. 
Our idea is to select a central tensor to reduce its dimension by one at each time, given the selection criterion that this truncation will lead to the least reconstruction error.
However, it is time-consuming to evaluate the reconstruction error of the original matrix. According to Eq.~\eqref{eq:epsilion-k}, we can utilize the error bound $\sqrt{\sum_{k=1}^{n-1} \epsilon_k^2}$ for a fast estimation of the yielded reconstruction error. In this case, only one $\epsilon_k$ changes, and it can be efficiently computed via the pre-computed eigenvalues.

\paratitle{Fast Performance Gap Computation}. At each time, we compute the performance gap before and after the dimension reduction ($d^{(l)} \rightarrow$ $d^{(l)}-1$) with the stop criterion. To obtain the performance $\tilde{p}$ after dimension reduction, we need to fine-tune the truncated model on the downstream task. We can also utilize the lightweight fine-tuning strategy in Section~\ref{sec-auxiliary} to  obtain $\tilde{p}$ by only tuning the auxiliary tensors. If the performance gap $\parallel p-\tilde{p}\parallel$ is smaller than a threshold $\Delta$ or the iteration number exceeds the predefined limit, the algorithm will end. Such an optimization algorithm is more stable to optimize  stacked architectures since it gradually reduces the dimension considering the reconstruction error and the performance gap. Actually, it is similar to the learning of variable matrix product states~\cite{iblisdir2007matrix} in physics, which optimizes the tensors one by one according to the sequence. As a comparison, our algorithm dynamically selects the matrix to truncate and is more suitable to PLMs. 

Algorithm~\ref{alg-squeezing_dimention} presents a complete procedure for our  algorithm. In practice, there are usually multiple parameter matrices to be optimized at each layer. 
This can be processed in a similar way: we select some matrices from one layer to optimize among all the considered matrices.

\begin{algorithm}
\small
    \caption{Training with dimension squeezing.}
    \begin{algorithmic}[1] 
        \Require: $L$ layers with corresponding central tensor $\Tensor{C}^{(l)}$ and  dimension $d^{(l)}$, threshold $\Delta$ and iteration step $iter$
        \State Evaluate loss ${p}$ = model($Inputs$)
        \State Perform MPO decomposition for each layer
        \For {$step = 1 \to iter$}
            \State Find the layer ($l^*$) with the least reconstruction error
            \State Compress MPO tensor by truncating $d^{(l^*)}$
            \State Fine-tuning auxiliary tensors with $\{\Tensor{C}^{(l)}\}_{l=1}^L$ fixed
            \State Evaluate loss $\tilde{p}$ = model($Inputs$)
            \If {$\parallel p-\tilde{p}\parallel > \Delta$}
                \State break
            \EndIf
        \EndFor
        \State \Return{Compressed model}
    \end{algorithmic}
\label{alg-squeezing_dimention}
\end{algorithm}

\subsection{Overall Compression Procedure}
Generally speaking, our approach can compress any PLMs with stacked architectures consisting of parameter matrices, even the compressed PLMs.  In other words, it can work with the existing PLM compression methods to further achieve a better compression performance. 
Here, we select ALBERT~\cite{lan2019albert} as a representative compressed PLM and apply our algorithm to ALBERT.

The procedure can be simply summarized as follows. 
First, we obtain the learned ALBERT model (complete) and perform the MPO-decomposition to the three major parameter matrices, namely word embedding matrix, self-attention matrix and feed-forward matrix\footnote{It introduces a parameter sharing mechanism to keep only one copy for both self-attention and feed-forward matrices.}. Each matrix will be decomposed into a central tensor and auxiliary tensors. Next, we perform the lightweight fine-tuning to update auxiliary tensors  until  convergence on downstream tasks. Then, we apply the dimension squeezing optimization algorithm to the three central tensors, \ie we select one matrix for truncation each time. After each truncation, we fine-tune the compressed model and further stabilize its performance.  This process will repeat until the performance gap or the iteration number  exceeds the pre-defined threshold. 

In this way, we expect that ALBERT can be further compressed. In particular, it can be fine-tuned in a more efficient way, with only a small amount of parameters to be updated. Section~\ref{sec:experimental-results} will demonstrate this.

\subsection{Discussion}\label{sec-discussion} 
In mathematics, MPO-based approximation can be considered as a special low-rank approximation method. Now, we compare it with other low-rank approximation methods, including SVD~\cite{henry19928}, CPD~\cite{hitchcock1927expression} and Tucker  decomposition~\cite{tucker1966some}.

We present the  categorization of these methods  in Table~\ref{tab-comparison}. For PLM compression, low-rank decomposition is only performed once, while it repeatedly performs forward propagation computation. Hence, we compare  their inference time complexities. 
Indeed, all the methods can be tensor-based decomposition (\ie a list of tensors for factorization) or matrix decomposition, and we characterize their time complexities with common parameters. Indeed, MPO and Tucker represent two categories of low-rank approximation methods. Generally, the algorithm capacity is larger with the increase of $n$ (more tensors). When $n>3$, MPO has smaller time complexity  than Tucker decomposition. 
It can be seen that SVD can be considered as a special case of MPO when tensor dimension $n=2$ and CPD is a special case of Tucker when the core tensor is the super-diagonal matrix.

\begin{table}[]
\small
\centering
\resizebox{\columnwidth}{!}{%
\begin{tabular}{cll}
\toprule[1pt]
Category & Method & Inference Time \\ \midrule[0.7pt]
\multirow{2}{*}{Tucker}    & Tucker$_{(d=1)}$(CP) &    $\mathcal{O}(nmd^2)$             \\
                           & Tucker$_{(d>1)}$     &    $\mathcal{O}(nmd+d^n)$             \\ \midrule[0.7pt]
\multirow{2}{*}{MPO}       & MPO$_{(n=2)}$(SVD)   &    $\mathcal{O}(2md^3)$           \\
                           & MPO$_{(n>2)}$        &    $\mathcal{O}(nmd^3)$             \\
\bottomrule[1pt]
\end{tabular}%
}
\caption{Inference time complexities of different low-rank approximation methods. Here, $n$ denotes the number of the tensors, $m$ denotes $\max(\{i_k\}_{k=1}^n)$ means the largest $i_k$ in input list, and $d$ denotes $\max(\{d'_k\}_{k=0}^n)$ means the largest dimension $d'_k$ in the truncated dimension list.}
\label{tab-comparison}
\end{table}

In practice, we do not need to strictly follow the original matrix size. Instead, it is easy to pad additional zero entries to enlarge matrix rows or columns, so that we can obtain different MPO decomposition results. It has demonstrated that different decomposition plans always lead to almost the same results~\cite{gao2020compressing}. In our experiments, we adopt an odd number of local tensors for MPO decomposition, \ie five local tensors (see supplementary materials). Note that MPO decomposition can work with other compression methods: it can further reduce the parameters from the matrices compressed by other methods, and meanwhile largely reduce the parameters to be fine-tuned.

\section{Experiments}
\label{sec:experiments}
In this section, we first set up the experiments, and then report the results and analysis.
\subsection{Experimental Setup}

\paratitle{Datasets}. We evaluate the effectiveness of compressing and fine-tuning PLMs of our approach MPOP on the General Language Understanding Evaluation~(GLUE) benchmark~\cite{wang2018glue}. GLUE is a collection of 9 datasets for evaluating natural language understanding systems. 
Following~\cite{sanh2019distilbert}, we report macro-score~(average of individual scores, which is slightly different from official GLUE score, since Spearman correlations are reported for STS-B and accuracy scores are reported for the other tasks) on the development sets for each task by fine-tuning MPOP.

\paratitle{Baselines}. Our baseline methods include: 

$\bullet$~\underline{BERT}~\cite{devlin2018bert}: The 12-layer BERT-base model was pre-trained on Wikipedia corpus released by Google.

$\bullet$~\underline{ALBERT}~\cite{lan2019albert}: It yields a highly compressed BERT variant with only 11.6M parameters, while maintains competitive performance, which serves as the major baseline.

$\bullet$~\underline{DistilBERT}~\cite{sanh2019distilbert}: It is trained via knowledge distillation with 6 layers. 

$\bullet$~\underline{MobileBERT}~\cite{sun2020mobilebert}:  It is equipped with bottleneck structures and a carefully designed balance between self-attentions and feed-forward networks.


All these models are released by Huggingface~\footnote{https://huggingface.co/}. We select these baselines because they are widely adopted and have a diverse coverage of compression techniques. 
Note that we do not directly compare our approach with other competitive methods~\cite{tambe2020edgebert} that require special 
optimization tricks or techniques (\eg hardware-level optimization). 

\paratitle{Implementation}. 
The original paper of ALBERT only reported the results of SST-2 and MNLI in GLUE. So we reproduce complete results denoted as ``ALBERT$_{\rm{rep}}$'' with the Huggingface implementation~\cite{wolf2020transformers}. 
Based on the pre-trained parameters provided by Huggingface, we also reproduce the results of 
BERT, DistilBERT and MobileBERT. To ensure a fair comparison, we adopt the same network architecture. For example, the number of self-attention heads, the hidden dimension of embedding vectors, and the max length of the input sentence are set to 12, 768 and 128, respectively. 
\begin{table*}
\small
\centering
\resizebox{\textwidth}{!}{
\begin{tabular}{llllllllllll} 
\toprule[1pt]
    \multicolumn{1}{c}{\multirow{1}{*}{\textbf{Experiments} }} & \multicolumn{1}{c}{\multirow{1}{*}{\textbf{Score} }}
         &
        \begin{tabular}[c]{@{}l@{}}\textbf{SST-2}\\\textbf{ (acc)} \end{tabular} & 
        \begin{tabular}[c]{@{}l@{}}\textbf{MNLI}\\\textbf{ (m\_cc)} \end{tabular} & \begin{tabular}[c]{@{}l@{}}\textbf{QNLI}\\\textbf{ (acc)} \end{tabular} & 
        \begin{tabular}[c]{@{}l@{}}\textbf{CoLA}\\\textbf{ (mcc)} \end{tabular} & \begin{tabular}[c]{@{}l@{}}\textbf{STS-B}\\\textbf{ ($\rho$)}\end{tabular} &
        \begin{tabular}[c]{@{}l@{}}\textbf{QQP}\\\textbf{ (acc)} \end{tabular} & \begin{tabular}[c]{@{}l@{}}\textbf{MRPC}\\\textbf{ (acc)} \end{tabular} & \begin{tabular}[c]{@{}l@{}}\textbf{RTE}\\\textbf{ (acc)} \end{tabular} & \begin{tabular}[c]{@{}l@{}}\textbf{WNLI}\\\textbf{ (acc)} \end{tabular} &  
        \begin{tabular}[c]{@{}l@{}}\textbf{Avg.}~\\\#Pr/\#To(M) \end{tabular} \\ 
    \midrule
        ALBERT$_{\rm{pub}}$ &-  & 90.3 & 81.6 & -  & - & - & - & - & - & - & 11.6/11.6 \\
        ALBERT$_{\rm{rep}}$~ & 78.9 & 90.6 & \textbf{84.5} & 89.4  & 53.4  & 88.2  &  89.1  & 88.5 & 71.1 & 54.9 & 11.6/11.6 \\
        MPOP & \textbf{79.7} &  \textbf{90.8} & 83.3 & \textbf{90.5} & \textbf{54.7} & \textbf{89.2} & \textbf{89.4} & \textbf{89.2} & \textbf{73.3}  & \textbf{56.3}  & \textbf{1.1/9} \\ 
    \midrule
        MPOP$_{\rm{full}}$ & 80.3 & 92.2 & 84.4 & 91.4 & 55.7 & 89.2 & 89.6 & 87.3 & 76.9 & 56.3 & 12.7/12.7  \\
        MPOP$_{\rm{full+LFA}}$ & 80.4 & 93.0  & 84.3  & 91.3 & 56.0 & 89.2 & 89.0 & 88.0  & 78.3 & 56.3 & 1.2/12.7 \\
        MPOP$_{\rm{dir}}$ & 68.6 & 86.6 & 79.2 & 81.9 & 15.0 & 82.5 & 87.0 & 74.3 & 54.2 & 56.3 &  1.1/9 \\
\bottomrule
\end{tabular}
}
\caption{Performance on GLUE benchmark obtained by fine-tuning ALBERT and MPOP. 
``ALBERT$_{\rm{pub}}$'' and 
``ALBERT$_{\rm{rep}}$'' denote the results from the original paper~\cite{lan2019albert} and reproduced by ours, respectively.
``\#Pr'' and ``\#To'' denote the number (in millions) of pre-trained parameters and total parameters, respectively. }
\label{tab:main_results}
\end{table*}

\subsection{Experimental Results}
\label{sec:experimental-results}
Note that our focus is to illustrate that our approach can improve either original~(uncompressed) or compressed PLMs. In our main experiments, we adopt ALBERT as the major baseline, and report the comparison results in Table~\ref{tab:main_results}. 

\ignore{
``ALBERT$_{\rm{pub}}$'' denotes that results are taken from original published papers~\cite{lan2019albert};  ``ALBERT$_{\rm{rep}}$'' denotes that results are reproduced by ours; ``MPOP$_{\rm{full}}$'' denotes training both auxiliary tensors and central tensors without reducing parameters; ``MPOP$_{\rm{full+LFA}}$'' denotes fine-tunig with auxiliary tensors as introduced in Section~\ref{sec-auxiliary}; ``MPOP$_{\rm{dir}}$ '' denotes fine-tuning after truncating the dimension of central tensors directly;``$\#$Pr/$\#$To'' denotes trainable parameters and total parameters respectively.
While, the comparisons with other baselines are presented in Section~\ref{sec-detail_analysis}.
}

\paratitle{Comparison with ALBERT}.
As shown in Table~\ref{tab:main_results}, our approach MPOP is very competitive in the GLUE benchmark, and it outperforms ALBERT in all tasks (except MNLI) with a higher overall score of 79.7. Looking at the last column, compared with ALBERT, MPOP reduces  total parameters by 22\% (\#To). In particular, it results in a significant reduction  of pre-trained parameters by 91\%~(\#Pr). Such a reduction is remarkable in lightweight fine-tuning, which dramatically improves the fine-tuning efficiency.  
By zooming in on specific tasks, the improvements over ALBERT are larger on CoLA, RTE and WNLI tasks. An interesting explanation is that RTE and WNLI tasks have small training sets (fewer than $4k$ samples). The lightweight fine-tuning strategy seems to work better with limited training data, which enhances the capacity of PLMs and prevents overfitting on downstream tasks. 

\paratitle{Ablation Results}. 
Our approach has incorporated two novel improvements: lightweight fine-tuning with auxiliary tensors and optimization with dimension squeezing. We continue to study their effect on the final performance. Here we consider three variants for comparison:
(1) MPOP$_{\rm{full}}$ and MPOP$_{\rm{full+LFA}}$ are full-rank MPO representation (without reconstruction error), and fine-tune \emph{all the tensors} and \emph{only auxiliary tensors}, respectively. This comparison is to examine whether only fine-tuning auxiliary tensors would lead to a performance decrease. 
(2) MPOP$_{\rm{dir}}$ directly optimizes the compressed model without the dimension squeezing algorithm. This variant is used to examine whether our optimization algorithm is more suitable for stacked architecture. 
Table~\ref{tab:main_results} (last three rows) shows the results when we ablate these.
In particular, the dimension squeezing algorithm plays a key role in improving our approach (a significant performance decrease for MPOP$_{\rm{dir}}$), since it is tailored to stacked architecture.
Comparing MPOP$_{\rm{full}}$ with MPOP$_{\rm{full+LFA}}$, it is noted that fine-tuning all the parameters seems to have a negative effect on performance. Compared with ALBERT, we speculate that fine-tuning a large model is more likely to overfit on small datasets (\eg RTE and MRPC). 
 
These results show that our approach is able to further compress ALBERT with fewer fine-tuning parameters. Especially, it is also helpful to improve the capacity and robustness of PLMs.   


\begin{table}[]
\small
\centering
\resizebox{\columnwidth}{!}{
    \begin{tabular}{lcccc} 
    \toprule[1pt]
        \multicolumn{1}{c}{\textbf{Models}} & 
        \begin{tabular}[c]{@{}c@{}}\textbf{WNLI}\\\textbf{ (acc)} \end{tabular} &  \begin{tabular}[c]{@{}l@{}}\textbf{MRPC}\\\textbf{ (acc)}\end{tabular} & \begin{tabular}[c]{@{}l@{}}\textbf{RTE}\\\textbf{ (acc)}\end{tabular} & 
        \begin{tabular}[c]{@{}c@{}}\textbf{Avg.}\\ \#Pr/\#To(M) \end{tabular} \\ 
        \midrule[0.7pt]
        BERT & 56.3 & \textbf{85.5} & 70.0 & 110/110 \\
        MPOP$_{\rm{B}}$ & 56.3 & 84.3 & \textbf{70.8} & \textbf{7.7/70.4}  \\ 
        \midrule[0.7pt]
        DistilBERT & 56.3 & 84.1 & 61.4 & 66/66 \\
        MPOP$_{\rm{D}}$ & 56.3 & \textbf{84.3} & \textbf{61.7} & \textbf{4.0/43.4} \\
        \midrule[0.7pt]
        MobileBERT & 56.2 & \textbf{86.0} & 63.5 & 25.3/25.3 \\
        MPOP$_{\rm{M}}$ & 56.2 & 85.3 & \textbf{65.7} & \textbf{4.4/15.4} \\
    \bottomrule[1pt]
\end{tabular}
}
\caption{Evaluation with different BERT variants.}
\label{tab-BERT-results}
\end{table}

\subsection{Detailed Analysis}
\label{sec-detail_analysis}
In this section, we perform a series of detailed analysis experiments for our approach.  

\paratitle{Evaluation with Other BERT Variants}. 
In general, our approach can be applied to either uncompressed or compressed PLMs. We have evaluated its performance with ALBERT. Now, we continue to test it with other BERT variants, namely original BERT,  DistilBERT and MobileBERT. The latter two BERT variants  are knowledge distillation based methods, and the distilled models can also be represented in the format of parameter matrix. We apply our approach to the three variants. 
Table~\ref{tab-BERT-results} presents the comparison of the three variants before and after the application of MPOP. 
As we can see, our approach can substantially reduce the network parameters, especially the parameters to be fine-tuned. Note that DistilBERT and MobileBERT are highly compressed models. These results show that our approach can further improve other compressed PLMs.

\paratitle{Evaluation on Different Fine-Tuning Strategies}.
Experiments have shown that our approach is able to largely reduce the number of parameters to be fine-tuned. Here we consider a more simple method to reduce the fine-tuning parameters, \ie only fine-tune the last layers of BERT. 
This experiment reuses the settings of BERT (12 layers) and our approach on BERT (\ie MPOP$_{\rm{B}}$  in Table~\ref{tab-BERT-results}).
We fine-tune the last 1-3 layers of BERT, and compare the performance with our approach MPOP$_{\rm{B}}$.  From Table~\ref{tab-fine-tune BERT}, we can see that such a simple way is much worse than our approach, especially on the RTE task. Our approach provides a more principled way for lightweight fine-tuning. By updating auxiliary tensors, it can better adapt to task-specific loss, and thus achieve better performance.

\begin{table}[]
\small
\centering
\begin{tabular}{cllll}
\toprule[1pt]
\textbf{Models} & \textbf{SST-2} & \textbf{MRPC} & \textbf{RTE} &
  \begin{tabular}[c]{@{}l@{}}\textbf{Avg.}\\ \textbf{\#Pr(M)}\end{tabular} \\ \midrule[0.7pt]
    BERT$_{10-12}$ & 91.9 & 76.5 & 67.2 & 45.7 \\
    BERT$_{11-12}$ & 91.7 & 75.3 & 62.8 & 38.6 \\ 
    BERT$_{12}$ & 91.4 & 72.1 & 61.4 & 31.5 \\ \midrule[0.7pt]
    MPOP$_{\rm{B}}$ & \textbf{92.6} & \textbf{84.3} & \textbf{70.8} & \textbf{10.1} \\ \bottomrule[1pt]
\end{tabular}
\caption{Comparison of different fine-tuning strategies on three GLUE tasks. The subscript number in BERT$_{(\cdot)}$ denotes the index of the layers to be fine-tuned. }
\label{tab-fine-tune BERT}
\end{table}

\begin{figure}
\centering
\subfigure[CPD \emph{v.s.} MPO.]{
\begin{minipage}[t]{0.49\columnwidth}
\centering
\includegraphics[width=\columnwidth]{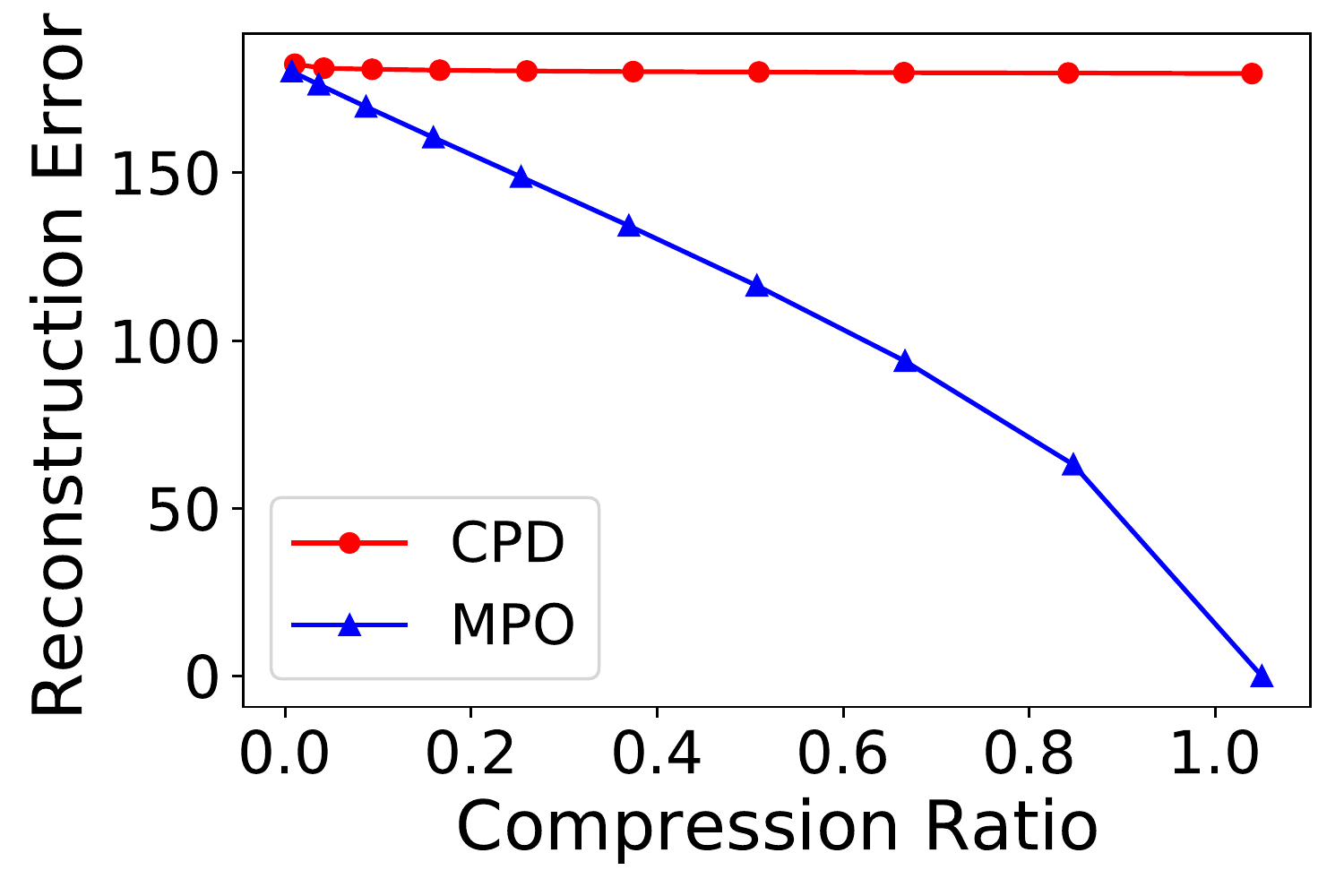} 
\end{minipage}%
}%
\subfigure[ \# of local tensors.]{
\begin{minipage}[t]{0.49\columnwidth}
\centering
\includegraphics[width=\columnwidth]{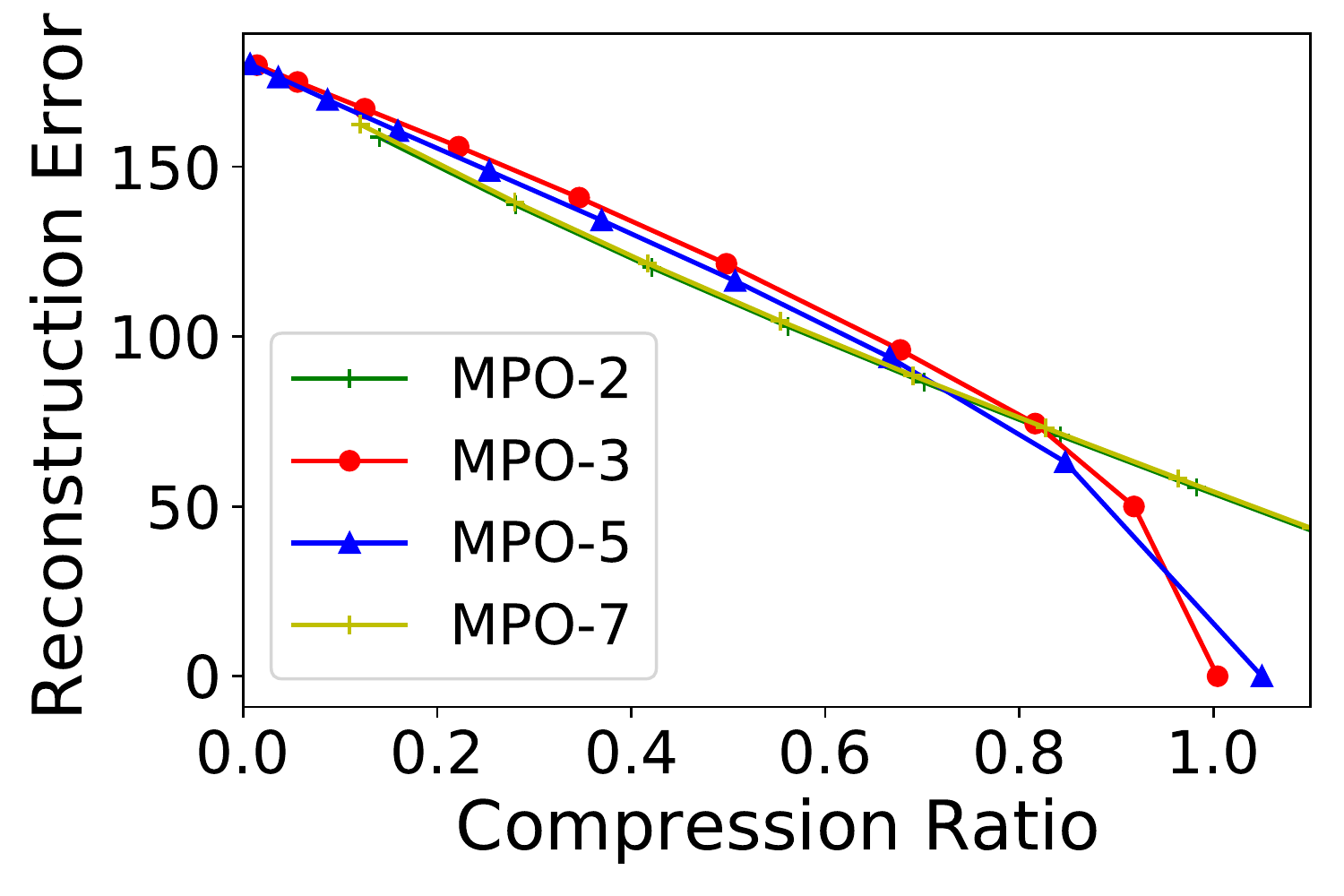} 
\end{minipage}%
}%
\caption{Comparison of different low-rank approximation variants.  $x$-axis denotes  the compression ratio~($\rho$ in Eq.~\eqref{eq:mpo-compression-rate}) and $y$-axis denotes the reconstruction error, measured in the Frobenius norm.}
\label{fig:fig2}
\end{figure}

\paratitle{Evaluation on Low-Rank Approximation}.
As introduced in Section~\ref{sec-discussion}, MPO is a special low-rank approximation method, and we first compare its compression capacity with other low-rank approximation methods. As shown in Table~\ref{tab-comparison}, MPO and Tucker decomposition represent two main categories of low-rank approximation methods. 
We select CPD~\cite{henry19928} for comparison because general Tucker decomposition~\cite{tucker1966some} cannot obtain results with reasonable memory. 
Our evaluation task is to compress the word embedding matrix of the released ``\emph{bert-base-uncased}'' model\footnote{https://huggingface.co/bert-base-uncased}. 
As shown in Figure~\ref{fig:fig2}(a), MPO achieves a smaller reconstruction error with all compression ratios, which shows that MPO is superior to CPD. 
Another hyper-parameter in our MPO decomposition is the number of local tensors ($n$).  
We further perform the same evaluation with different numbers of local tensors ($n=3,5,7$). 
From Figure~\ref{fig:fig2}(b), it can be observed that our method is relatively stable with respect to the number of local tensors. 
Overall, a larger $n$ requires a higher time complexity and can yield flexible decomposition. Thus, we set $n=5$ for making a trade-off between flexibility and efficiency. 

%

\section{Conclusion}

We proposed an MPO-based PLM compression method. With MPO decomposition, we were able to reorganize and aggregate information in central tensors effectively. Inspired by this, we designed a novel fine-tuning strategy that only needs to fine-tune the parameters in auxiliary tensors. We also developed a dimension squeezing training algorithm for optimizing low-rank approximation over stacked network architectures. Extensive experiments had demonstrated the effectiveness of our approach, especially on the reduction of fine-tuning parameters. We also empirically found that such a fine-tuning way was more robust to generalize on small training datasets.  To our knowledge, it is the first time that MPO decomposition had been applied to compress PLMs. In future work, we will consider exploring more decomposition structures for MPO. 

\section*{Acknowledgments}
This work was partially supported by the National Natural Science Foundation of China under Grants No. 61872369, 61832017 and 11934020, Beijing Academy of Artificial Intelligence (BAAI) under Grant No. BAAI2020ZJ0301, Beijing Outstanding Young Scientist Program under Grant No. BJJWZYJH012019100020098, the Fundamental Research Funds for the Central Universities and the Research Funds of Renmin University of China under Grant No. 18XNLG22, 19XNQ047, 20XNLG19 and 21XNH027. Xin Zhao and Zhong-Yi Lu are the corresponding authors.

\bibliographystyle{acl_natbib}
\bibliography{acl2021}


\end{document}


\maketitle

\appendix

We offer some proof as supplementary materials to help authors better understand our model. The appendix includes \underline{3 pages} and is organized into sections:
\begin{itemize}
    \item Tensor and Matrix Product Operators
    \item Theorem
    \item Experiment
\end{itemize}

\section{Tensor and Matrix Product Operators}
As introduced in~\cite{cichocki2009nonnegative}, the concept of tensor is specified as:

\paratitle{Definition1} \\
(Tensor). Let $D_1,D_2...,D_N \in N$ denote index upper bounds. A tensor $\Tensor{T}\in \mathbb{R}^{D_1,...,D_n}$ of order $N$ is an $N$-way array where elements $\Tensor{T}_{d_1,d_2,...,d_n}$ are indexed by $d_n\in\{1,2,...,D_n\}$ for $1\leq n\leq N$ 

\paratitle{Definition2} \\
(Matrix product operator).
We can reshape a matrix to high order tensor, denote as:
\begin{equation}
    \Matrix{M}_{x\times y} = \Matrix{M}_{i_1i_2...i_n,j_1j_2...j_n}
\end{equation}
Here, the one-dimensional coordinate $x$ of the input signal $\mathbf{x}$ with dimension $N_x$ is reshaped into a coordinate in a $n$-dimensional space, labelled by $(i_1 i_2 \cdots i_n)$.
Hence, there is a one-to-one mapping between $x$ and $(i_1 i_2 \cdots i_n)$.
Similarly, the one-dimensional coordinate $y$ of the output signal $\mathbf{y}$ with dimension $N_y$ is also reshaped into a coordinate in a $n$-dimensional space, and there is a one-to-one correspondence between $y$ and $(j_1j_2\cdots j_n)$.
If $I_k$ and $J_k$ are the dimensions of $i_k$ and $j_k$, respectively, then
\begin{equation}
    \prod_{k=1}^{n}I_{k} = N_{x}, \quad \prod_{k=1}^{n}J_{k} = N_{y}  . \label{eq:dims}
\end{equation}

The MPO representation of $M$ is obtained by factorizing it into a product of $n$ local tensors
\begin{eqnarray}
 M_{i_1\cdots i_n,j_1\cdots j_n} =  \Tensor{T}^{(1)} [i_1,j_1] \cdots \Tensor{T}^{(n)} [i_n,j_n]  \label{Eq:MPO}
\end{eqnarray}
where $\Tensor{T}^{(k)}[j_k,i_k]$ is a $D_{k-1}\times D_{k}$ matrix with $D_k$ the virtual basis dimension on the bond linking $\Tensor{T}^{(k)}$ and $\Tensor{T}^{(k+1)}$ with $D_0=D_n=1$.

\section{Theorem}
\newtheorem{thm}{\bf Theorem}
\begin{thm}\label{thm1}
Suppose that the tensor $\textbf{W}^{(k)}$ of matrix $W$ that is satisfy 
\begin{align}
    &\Matrix{W} = \Matrix{W}^{(k)} + \Matrix{E}^{(k)}  ,D( \textbf{W}^{(k)}) = d_k,\notag\\
    &where\quad||\Matrix{E}^{(k)}||_F^2 = \epsilon_k^2 , k = 1,...,d-1. 
\end{align}
Then $MPO~(\Matrix{W})$ with the $k$-th bond dimension $d_k$ upper bound of truncation error satisfy:
\begin{equation}
    ||\Matrix{W}-MPO~(\Matrix{W})||_F \leq \sqrt{\sum_{k=1}^{d-1} \epsilon_k^2}
\end{equation}
\end{thm}

$Proof.$ The proof is by induction. For $n = 2$ the statement follows from the properites of the SVD. Consider an arbitrary $n > 2$. Then the first unfolding $\Matrix{W}^{(1)}$ is decomposed as
\begin{equation}
    \Matrix{W}^{(1)} = \Matrix{U}_1 \Matrix{\lambda}_1 \Matrix{V}_1 + \Matrix{E}^{(1)} = \Matrix{U}_1\Matrix{B}^{(1)} + \Matrix{E}^{(1)}
\end{equation}
where $\Matrix{U}_1$ is of size $r_1\times i_1 \times j_1$ and $|| \Matrix{E}^{(1)}||_F^2 = \epsilon_1^2$. The matrix $\Matrix{B}_1$ is naturally associated with a $(n-1)$-dimensional tensor $\Tensor{B}^{(1)}$ with elements $\Tensor{B}^{(1)}(\alpha , i_2,j_2, ..., i_n,j_n)$, which will be decomposed further. This means that $\Matrix{B}_1$ will be approximated by some other matrix $\hat{\Matrix{B}_1}$. From the properties of the SVD it follows that $\Matrix{U}_1^{T}\Matrix{E}^{(1)}=0$, and thus

\begin{align}
    & ||\Matrix{W}-\Tensor{B}^{(1)}||^2_F \notag\\ 
    & = ||\Matrix{W}_1 - \Matrix{U}_1\hat{\Matrix{B}_1}||_F^{2} \notag\\
    & = ||\Matrix{W}_1 - \Matrix{U}_1(\hat{\Matrix{B}_1} + \Matrix{B}_1 - \Matrix{B}_1)||_F^{2}\notag\\
    & = ||\Matrix{W}_1 - \Matrix{U}_1\Matrix{B}_1||_F^{2} + ||\Matrix{U}_1(\hat{\Matrix{B}_1} - \Matrix{B}_1)||_F^2
\end{align}

and since $\Matrix{U}_1$ has orthonormal columns,
\begin{equation}
    ||\Matrix{W}-\Tensor{B}^{(1)}||_F^2 \leq \epsilon_1^2 + ||\Matrix{B}_1-\hat{\Matrix{B}_1}||_F^2.
    \label{eq:thm1}
\end{equation}
and thus it is not difficult to see from the orthonormality of columns of $\Matrix{U}_1$ that the distance of the $k$-th unfolding $(k=2,...,d_k-1)$ of the $(d-1)$-dimensional tensor $\Tensor{B}^{(1)}$ to the $d_k$-th rank matrix cannot be larger then $\epsilon_k$. Proceeding by induction, we have 
\begin{equation}
    ||\Matrix{B}_1 - \hat{\Matrix{B}_1} ||_F^2 \leq \sum_{k=2}^{d-1} \epsilon_k^2,
\end{equation}
combine with Eq.~\eqref{eq:thm1}, this complets the proof.

\section{Experiment}
\subsection{Additional Details of MPO}
\label{sec:additional-details}
In this paper, the MPOP is proposed for compressing pre-trained Language Models. In order to show that the process of incorporating several MPO sturctures into ALBERT-based  and  BERT-based pre-trained language models respectively. We introduce MPO decomposition in ALBERT and BERT details as follows:

\begin{table}[h]
\small
\centering
\resizebox{\linewidth}{!}{
\begin{tabular}{lll}
\toprule[1pt]
Layers                          & Matrix shape               & \begin{tabular}[c]{@{}l@{}}\multirow{1}{*}{MPO shape} \\ {[$d_{k-1},i_k,j_k,d_k$]}\end{tabular} \\ \midrule[0.7pt]
\multirow{5}{*}{AlbertEmbeddings} & \multirow{5}{*}{$30000\times128$} & $\Tensor{A}_1$:$[1,5,2,10]$                \\
                                &                            & $\Tensor{A}_2$:$[10,10,2,200]$             \\
                                &                            & $\Tensor{C}$:$[200,10,4,480]$             \\
                                &                            & $\Tensor{A}_3$:$[480,10,4,12]$             \\
                                &                            & $\Tensor{A}_4$:$[12,6,2,1]$                \\ \midrule[0.7pt]
\multirow{10}{*}{AlbertLayer}  & \multirow{5}{*}{$768\times3072$}  & $\Tensor{A}_1$:$[1,3,4,12]$                \\
                                &                            & $\Tensor{A}_2$:$[12,4,4,192]$              \\
                                &                            & $\Tensor{C}$:$[192,4,8,384]$              \\
                                &                            & $\Tensor{A}_3$:$[384,4,6,16]$              \\
                                &                            & $\Tensor{A}_4$:$[16,4,4,1]$                \\\cline{2-3}
                                & \multirow{5}{*}{$3072\times768$}  & $\Tensor{A}_1$:$[1,4,3,12]$                \\
                                &                            & $\Tensor{A}_2$:$[12,4,4,192]$              \\
                                &                            & $\Tensor{C}$:$[192,8,4,384]$              \\
                                &                            & $\Tensor{A}_3$:$[384,6,4,16]$              \\
                                &                            & $\Tensor{A}_4$:$[16,4,4,1]$                \\ \midrule[0.7pt]
\multirow{5}{*}{\begin{tabular}[c]{@{}l@{}}AlbertAttention\\ (query/key/value/\\output)\end{tabular}} & \multirow{5}{*}{$768\times768$} & $\Tensor{A}_1$:$[1,3,4,12]$ \\
                                &                            & $\Tensor{A}_2$:$[12,4,4,192]$              \\
                                &                            & $\Tensor{C}$:$[192,4,4,192]$              \\
                                &                            & $\Tensor{A}_3$:$[192,4,4,12]$              \\
                                &                            & $\Tensor{A}_4$:$[12,4,3,1]$                \\ \bottomrule[1pt]
\end{tabular}%
}
\caption{ALBERT MPO Decomposition Shape}
\end{table}
There is some slight difference of MPO structure between ALBERT and BERT. In word embedding layer, we use MPO to decompose a matrix of shape [30720,768] rather than [30522,768], for ``30522'' can not be reshaped to dimensions of $i_k$ as introduced in Eq.~\eqref{eq:dims}. Specifically, We get [30720,768] by zero padding first, then we apply MPO decomposition, at last, we clip the paddings before computing with input tokens. In intermediate and output layers, BERT and ALBERT share all of the shape of matrix.  

\begin{table}[h]
\small
\centering
\resizebox{\linewidth}{!}{
\begin{tabular}{lll}
\toprule[1pt]
Layers                          & Matrix shape               & \begin{tabular}[c]{@{}l@{}}\multirow{1}{*}{MPO shape} \\ {[$d_{k-1},i_k,j_k,d_k$]}\end{tabular} \\ \midrule[0.7pt]
\multirow{5}{*}{BertEmbeddings} &    \multirow{5}{*}{$30720\times768$} & $\Tensor{A}_1:[1,5,2,10]$             \\
                                &                            & $\Tensor{A}_2:[10,10,2,200]$             \\
                                &                            & $\Tensor{C}:[200,10,4,480]$             \\
                                &                            & $\Tensor{A}_3:[480,10,4,12]$             \\
                                &                            & $\Tensor{A}_4:[12,6,2,1]$                \\ \midrule[0.7pt]
\multirow{5}{*}{BertIntermediate}  & \multirow{5}{*}{$768\times3072$}  & $\Tensor{A}_1:[1,3,4,12]$                \\
                                &                            & $\Tensor{A}_1:[12,4,4,192]$              \\
                                &                            & $\Tensor{C}$:$[192,4,8,384]$              \\
                                &                            & $\Tensor{A}_3$:$[384,4,6,16]$              \\
                                &                            & $\Tensor{A}_4$:$[16,4,4,1]$                \\ \midrule[0.7pt]
\multirow{5}{*}{Bertoutput}         & \multirow{5}{*}{$3072\times768$}  & $\Tensor{A}_1$:$[1,4,3,12]$                \\
                                &                            & $\Tensor{A}_2$:$[12,4,4,192]$              \\
                                &                            & $\Tensor{C}$:$[192,8,4,384]$              \\
                                &                            & $\Tensor{A}_3$:$[384,6,4,16]$              \\
                                &                            & $\Tensor{A}_4$:$[16,4,4,1]$                \\ \midrule[0.7pt]
\multirow{5}{*}{\begin{tabular}[c]{@{}l@{}}BertAttention\\ (query/key/value/\\output)\end{tabular}} & \multirow{5}{*}{$768\times768$} & $\Tensor{A}_1$:$[1,3,4,12]$ \\
                                &                            & $\Tensor{A}_2$:$[12,4,4,192]$              \\
                                &                            & $\Tensor{C}$:$[192,4,4,192]$              \\
                                &                            & $\Tensor{A}_3$:$[192,4,4,12]$              \\
                                &                            & $\Tensor{A}_4$:$[12,4,3,1]$                \\ \bottomrule[1pt]
\end{tabular}%
}
\caption{BERT MPO Decomposition Shape}
\end{table}

\subsection{Experimental Details in Pre-trained Language Modeling}
Now, we report some details of experiments as a relevant supplementary material. 
Firstly,  we expand all the matrices $\{\Matrix{M}_k\}_{k=1}^{N}$ in ALBERT into MPO structure with $\big\{\{\Tensor{A}_1, \Tensor{A}_2,\Tensor{C},\Tensor{A}_3,\Tensor{A}_4\}_k\big\}_{k=1}^{N}$. Specific details in~\ref{sec:additional-details}. In the experimental of main text, "MPOP$_{\rm{full}}$" means that we fine-tune all these tensors compare with "MPOP$_{\rm{full+LFA}}$"denotes that we fine-tune these tensors with central tensor fixed. 
Then, we can further compressing the MPO structure by truncating $\{d_k\}$ to $\{d'_k\}$ as described in the main text. At the same time, Dimension-Squeezing method can also be used for compression and fine-tuning.

\paratitle{Hardware}
We trained our model on one machine with 4 NVIDIA Titan V GPUs. For our base models, we adopt all these models released by Huggingface~\footnote{https://huggingface.co/}. 

\paratitle{Optimizer}
We used the Adam optimizer and vary the learning rate over the course of training. The vary
formula~\cite{vaswani2017attention} is follows in our work. We also used the $warmup\_steps = 1000$.

\bibliographystyle{acl_natbib}
\bibliography{appendix}